\documentclass{article}

\PassOptionsToPackage{numbers, compress}{natbib}

\usepackage[preprint]{neurips_2019}
\usepackage{neurips_2019}

\usepackage[utf8]{inputenc} %
\usepackage[T1]{fontenc}    %
\usepackage{hyperref}       %
\usepackage{url}            %
\usepackage{booktabs}       %
\usepackage{amsfonts}       %
\usepackage{nicefrac}       %
\usepackage{microtype}      %
\usepackage{subfigure}
\usepackage{graphicx}
\usepackage{amsmath}
\usepackage{xcolor}
\usepackage[export]{adjustbox}

\title{Hamiltonian Graph Networks with ODE Integrators}

\author{%
  Alvaro Sanchez-Gonzalez \\
  DeepMind \\
  London, UK \\
  \texttt{alvarosg@google.com} \\
  \And
  Victor Bapst \\
  DeepMind \\
  London, UK \\
  \texttt{vbapst@google.com} \\
  \And
  Kyle Cranmer \\
  NYU \\
  New York, USA \\
  \texttt{kc90@nyu.edu} \\
  \And
  Peter Battaglia \\
  DeepMind \\
  London, UK \\
  \texttt{peterbattaglia@google.com}
}

\newcommand{\deltamodel}[0]{DeltaGN}
\newcommand{\odemodel}[0]{OGN}
\newcommand{\hamiltonianmodel}[0]{HOGN}
\newcommand{\truehamiltonian}[0]{true Hamiltonian}
\newcommand{\captionletter}[1]{\textbf{(#1)}}

\begin{document}

\maketitle

\begin{abstract}
We introduce an approach for imposing physically informed inductive biases in learned simulation models. We combine graph networks with a differentiable ordinary differential equation integrator as a mechanism for predicting future states, and a Hamiltonian as an internal representation. We find that our approach outperforms baselines without these biases in terms of predictive accuracy, energy accuracy, and zero-shot generalization to time-step sizes and integrator orders not experienced during training. This advances the state-of-the-art of learned simulation, and in principle is applicable beyond physical domains.

\end{abstract}

\section{Introduction}
Learning to simulate complex physical processes has been shown to benefit from rich inductive biases about the structure of the system's state, such as modeling particles and their interactions via nodes and edges in a graph \cite{battaglia2016interaction,chang2017compositional,sanchezgonzalez2018graph,mrowca2018flexible,li2018learning}, and optimizing internal representations to predict known physical quantities \cite{seo2019differentiable}. Here we explore another class of physically informed inductive biases: that a system's dynamics can be modeled as an ordinary differential equation (ODE) and formulated as Hamiltonian mechanics. These have also been individually studied recently by \citet{chen2018neural,rubanova2019latent} (ODE bias), and \citet{greydanus2019hamiltonian} (Hamiltonian bias), but here we combine them and incorporate both into graph-network-based neural architectures \cite{battaglia2018relational,battaglia2016interaction,sanchezgonzalez2018graph} for learning simulation.

Our experiments show that our trained models have greater predictive accuracy than baselines, as well as better energy conservation (via the Hamiltonian inductive bias) and stronger generalization to novel time-steps and integrator orders (via the ODE integrator inductive bias). These inductive biases also carry trade-offs: with lower order integrators and coarse time-steps, both our learned Hamiltonian model as well as a model which uses the true Hamiltonian derived from physics struggle, because the Hamiltonian puts hard constraints on how the higher order structure of the dynamics must be modeled, whereas the less constrained models can learn ways of approximating such structure.

\section{Background}\label{sec:background}

\paragraph{Graph networks}

We represent a particle system as a graph whose nodes correspond to particles, and with edges connecting all nodes to each other. All of our models use a graph network (GN) \cite{battaglia2018relational}, which operates on graphs $G = (\mathbf{u}, V, E)$ with global features, $\mathbf{u}$, and variable numbers of nodes, $V$, and edges, $E$. Here we focus on GNs which can compute per-node outputs, $V' = \mathrm{GN}_V(G)$, and global outputs $
\mathbf{u}' = \mathrm{GN}_\mathbf{u}(G)$. Preliminary experiments show our GN model outperformed an MLP-based approach by several orders of magnitude, consistently with previous work \cite{battaglia2016interaction,sanchezgonzalez2018graph}.

\paragraph{Numerical integrators for solving ODEs}

Given a first-order ODE and initial conditions, numerical integration can be used to approximate solutions to the initial value problem,
\begin{equation}
\label{eq:first_order_ode}
\mathbf{y} \equiv \mathbf{y}(t)
\quad,\quad
\mathbf{\dot y} \equiv \frac{\mathrm{d}\mathbf{y}}{\mathrm{d}t} = f_{\mathbf{\dot y}}(t, \mathbf{y})
\quad,\quad 
\mathbf{y}(t_0) = \mathbf{y}_{0} \quad .
\end{equation}
The family of Runge-Kutta (RK) integrators are fully differentiable and can generate trajectories via iterations of the form, $\mathbf{y}_{n+1}~=~\mathrm{RK}(t_n, \Delta t, \mathbf{y}_{n}, f_{\mathbf{\dot y}})$, where $\Delta t = t_{n+1}-t_{n}$. 
The lowest order RK integrator (Euler integrator, RK1) iterates as, $\mathbf{y}_{n+1} = \mathbf{y}_n + \Delta t\cdot f_{\mathbf{\dot y}}(t_n,\mathbf{y}_n)$. Higher order RK integrators produce more accurate trajectories by composing multiple queries to the function $f_{\mathbf{\dot y}}$. Here we explore first- through fourth-order RK integrators: RK1, RK2, RK3, and RK4 (See Supp. Sec. \ref{sec:symplectic integrators} for results with symplectic integrators).

\paragraph{Hamiltonian mechanics}

In Hamiltonian mechanics, the Hamiltonian, $\mathcal{H}(\mathbf{q}, \mathbf{p})$, is a function of the canonical position, $\mathbf{q}$, and momentum, $\mathbf{p}$, coordinates, and usually corresponds to the energy of the system
\footnote{Our methods are also compatible with time-dependent Hamiltonians, $\mathcal{H}(t, \mathbf{q}, \mathbf{p})$. However, since our dataset does not require time dependency, for simplicity we will omit $t$ everywhere from here on.}. 
The dynamics of the system follow Hamilton's equations, two first-order ODEs (analogous to Eq. \ref{eq:first_order_ode} with $\mathbf{y} = (\mathbf{q},\mathbf{p})$),
\begin{equation} \label{eq:hamilton_equations}
\mathbf{\dot q}\equiv\frac{\mathrm{d}\mathbf{q}}{\mathrm{d}t} = \frac{\partial \mathcal{H}}{\partial \mathbf{p}}
\quad,\quad
\mathbf{\dot p}\equiv\frac{\mathrm{d}\mathbf{p}}{\mathrm{d}t} = -\frac{\partial \mathcal{H}}{\partial \mathbf{q}}
\quad\Rightarrow\quad
(\mathbf{\dot q},\mathbf{\dot p})\equiv\frac{\mathrm{d}(\mathbf{q},\mathbf{p})}{\mathrm{d}t} = f_{\mathbf{\dot q}, \mathbf{\dot p}}(\mathbf{q}, \mathbf{p})\quad .
\end{equation}

\section{Models}

\begin{figure*}[t!]
\begin{center}
    \adjincludegraphics[width=1.0\textwidth]{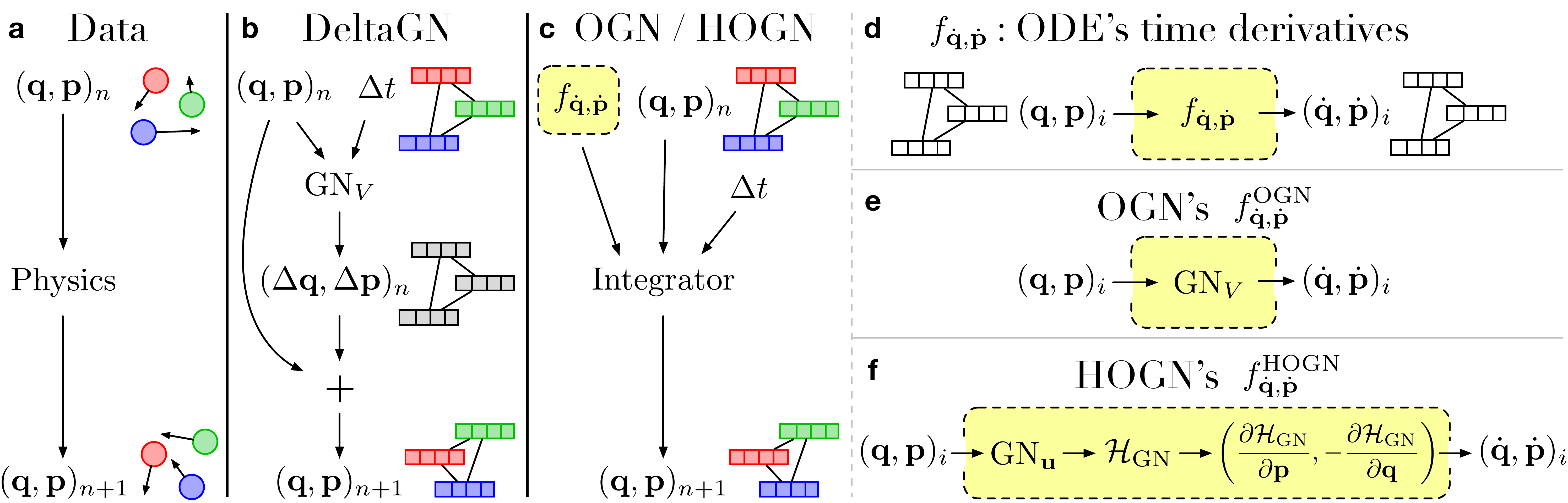}
    \caption{\captionletter{a} The data reflects a system's temporal dynamics, which is governed by physics. \captionletter{b} The baseline \deltamodel{} model takes as input a state represented by a graph and a time-step, and uses a $\mathrm{GN}_V$ to predict state changes. \captionletter{c} Our \odemodel{} and \hamiltonianmodel{} models take as input an input state, time-step, and a function to evaluate the ODE's time derivatives, $f_{\dot{\mathbf{q}}, \dot{\mathbf{p}}}$, and uses an integrator, which queries $f_{\dot{\mathbf{q}}, \dot{\mathbf{p}}}$ at different points, to predict the state after the time-step. \captionletter{d} The $f_{\dot{\mathbf{q}}, \dot{\mathbf{p}}}$ takes as input any state and outputs its time derivatives. \captionletter{e} The \odemodel{} model uses a $\mathrm{GN}_V$ as $f^{\textrm{OGN}}_{\dot{\mathbf{q}}, \dot{\mathbf{p}}}$. \captionletter{f} The \hamiltonianmodel{} model's $f^{\textrm{HOGN}}_{\dot{\mathbf{q}}, \dot{\mathbf{p}}}$ uses a $\mathrm{GN}_\mathbf{u}$ to predict the Hamiltonian, which is then differentiated w.r.t. the input state.}
    \label{fig:schematic}
\end{center}
\end{figure*}

\paragraph{Delta graph network (\deltamodel{})} 
Our main baseline, a ``delta graph network'' (\deltamodel{}), replicates previous approaches for learning to simulate \cite{battaglia2016interaction,sanchezgonzalez2018graph}, and directly predicts changes to $\mathbf{q}$ and $\mathbf{p}$,
\begin{align}
(\mathbf{q}, \mathbf{p})_{n+1} &= 
(\mathbf{q}, \mathbf{p})_{n} + 
(\Delta \mathbf{q}, \Delta \mathbf{p})_{n} \quad \text{, where} \quad (\Delta \mathbf{q}, \Delta \mathbf{p})_{n} \leftarrow \mathrm{GN}_{V}(\Delta t, \mathbf{q}_n, \mathbf{p}_n,  \mathbf{c}; \phi) \label{eq:delta_model} \quad .
\end{align}
The $\mathbf{c}$ are static parameters (masses, spring constants) of the system, and $\phi$ are the neural network parameters. The $\mathrm{GN}$'s signature matches the integrator's, and so is analogous to learning an integrator.

\paragraph{ODE graph network (\odemodel{})}

Our ``ODE graph network'' (\odemodel{}) imposes an ODE integrator as an inductive bias in the GN, by assuming that the dynamics of $(\mathbf{q}, \mathbf{p})$ follow a first-order ODE (Eq. \ref{eq:first_order_ode}). We train a neural network that learns the ODE, that is, learns to produce the time derivatives $(\mathbf{\dot q}, \mathbf{\dot p})$ (which are independent from $\Delta t$). We again use the node output of a GN to model the per-particle time derivatives, and provide the GN, together with the initial conditions and $\Delta t$, to an RK integrator, 
\begin{align}
(\mathbf{q}, \mathbf{p})_{n+1} &= \mathrm{RK}\left(\Delta t, (\mathbf{q}, \mathbf{p})_{n}, f^{\textrm{OGN}}_{\dot{\mathbf{q}}, \dot{\mathbf{p}}}\right) \label{eq:model_integrator} \\
f^{\textrm{OGN}}_{\dot{\mathbf{q}}, \dot{\mathbf{p}}}(\mathbf{q}, \mathbf{p}) &\equiv \mathrm{GN}_{V}(\mathbf{q}, \mathbf{p}, \mathbf{c}; \phi) = \left(\mathbf{\dot q},\mathbf{\dot p}\right) \quad . 
\end{align}
The $f_{\dot{\mathbf{q}}, \dot{\mathbf{p}}}$ is a function that the integrator can use to operate on any $(\mathbf{q}, \mathbf{p})$ and query more than once. For fixed $\Delta t$ and when using the RK1/Euler integrator (see Sec.~\ref{sec:background}) this is equivalent (up to a scale factor) to the \deltamodel{}.

\paragraph{Hamiltonian ODE graph network (\hamiltonianmodel{})}

Our ``Hamiltonian ODE graph network'' (\hamiltonianmodel{}) imposes further constraints by using a $\textrm{GN}_\mathbf{u}$ to compute a single scalar for the system---the Hamiltonian---via the global output, analytically differentiating it with respect to its inputs, $\mathbf{q}$ and $\mathbf{p}$, and, in accordance with Hamilton's equations (Eq. \ref{eq:hamilton_equations}), treating these gradients as $-\mathbf{\dot p}$ and $\mathbf{\dot q}$, respectively,
\begin{align}
\mathcal{H}_{\mathrm{GN}}(\mathbf{q}, \mathbf{p}) &= \mathrm{GN}_\mathbf{u}(\mathbf{q}, \mathbf{p}, \mathbf{c}; \phi) \\ 
f^{\textrm{HOGN}}_{\dot{\mathbf{q}}, \dot{\mathbf{p}}}(\mathbf{q}, \mathbf{p})
&\equiv \left(\frac{\partial \mathcal{H}_{\mathrm{GN}}}{\partial \mathbf{p}}
,-\frac{\partial \mathcal{H}_{\mathrm{GN}}}{\partial \mathbf{q}}\right) = \left(\mathbf{\dot q}, \mathbf{\dot p}\right) \label{eq:network_output_derivatives}
 \quad .
\end{align}
The resulting ODE defined by $f^{\textrm{HOGN}}_{\dot{\mathbf{q}}, \dot{\mathbf{p}}}$ can be integrated similarly to the \odemodel{} by passing those functions to the integrator (Eq. \ref{eq:model_integrator}). Crucially, the $\mathcal{H}_{\mathrm{GN}}$ is not supervised directly, but instead learned end-to-end through the integrator.

\section{Results}

\begin{figure*}[t!]
\begin{center}
    \adjincludegraphics[height=0.13\textheight, trim={{0.01\width} 0 {.01\width} 0},clip=true]{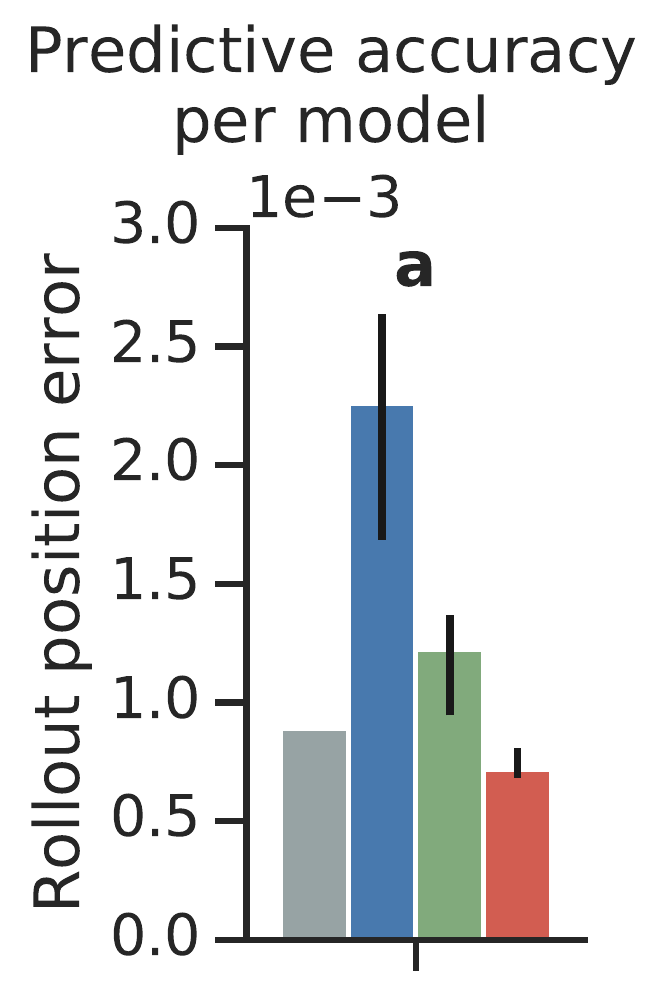}
    \adjincludegraphics[height=0.13\textheight, trim={{0.01\width} 0  {.01\width} 0},clip=true]{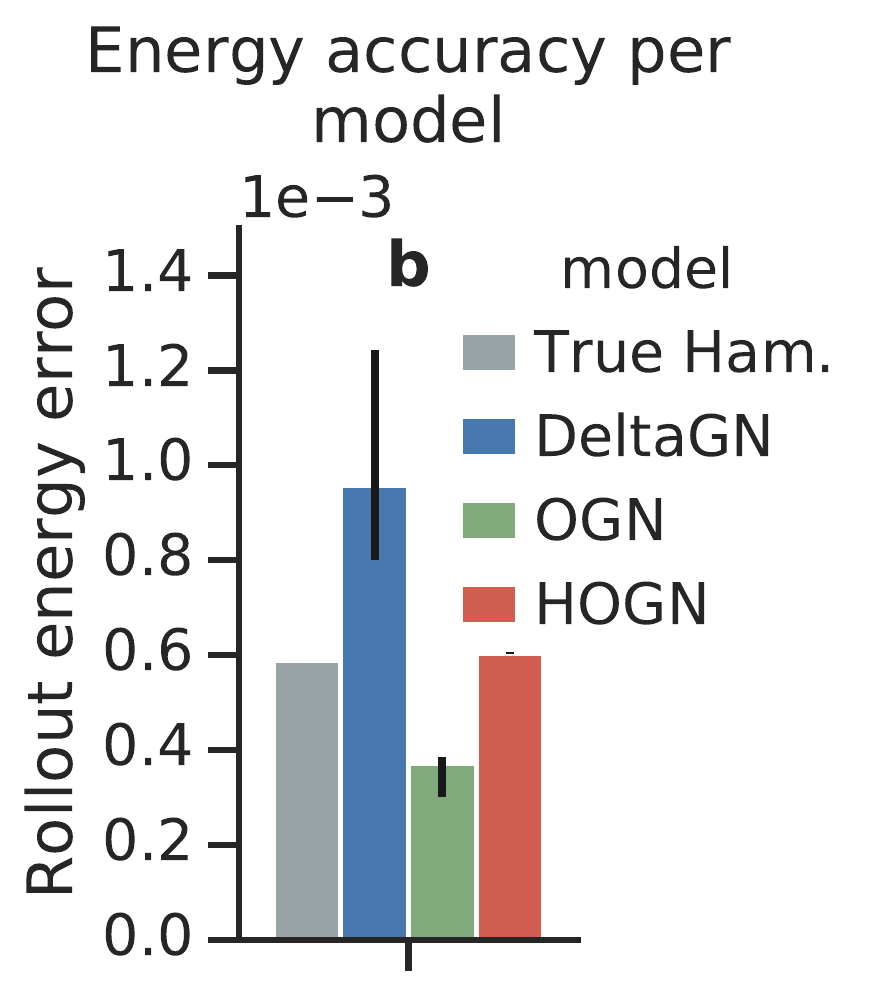}
    \raisebox{0.0\textheight}{\adjincludegraphics[height=0.13\textheight, trim={{0.01\width} 20  {.01\width} 0},clip=true]{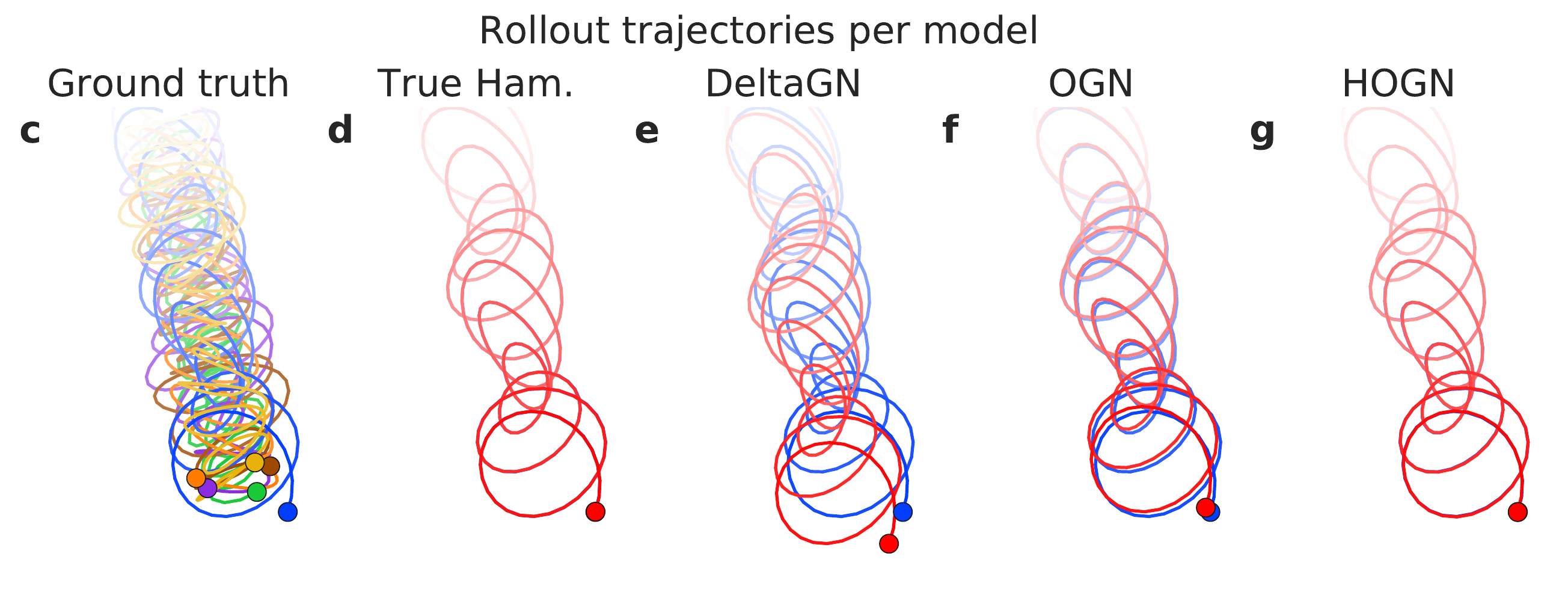}}
    \caption{\captionletter{a} Predictive accuracy (on 20-step trajectory) across models using RK4 for the ODE based models. The \hamiltonianmodel{} is most accurate. \captionletter{b} Energy accuracy across the same models. \captionletter{c-g} Last 300 steps of a 500-step trajectory of a 6-particle system, where dots indicate the final position of the particles (colors fade into the past). \captionletter{c} Ground truth trajectory for all particles. \captionletter{d-g} Trajectory for one of the particles (blue) superimposed by the trajectory obtained (red) when integrating the True Hamiltonian, or \captionletter{e-g} using the best seed of the different learned models, at a time step of 0.1. While errors of all models are very small at the beginning of the trajectory, the \hamiltonianmodel{} is the only learned model still indistinguishable from ground truth at the end of the long 500-step trajectory (\href{https://tinyurl.com/hogn-example-trajectories}{\textcolor{blue}{\underline{video link}}}).}
    \label{fig:overview}
\end{center}
\end{figure*}

We train and test our approach on datasets consisting of particle systems where particle $j$ exerts a spring force on particle $i$, as defined by Hooke's law, $\mathbf{F}^{ij} = -k^{ij}\cdot(\mathbf{q}^i - \mathbf{q}^j)$,
where $k^{ij}$ is the spring constant. All systems contained between $4$ and $9$ particles, and simulations were computed with time-steps of $0.005$. To form the data from the raw simulations, we sub-sampled the trajectories at a fixed time-step of $0.1$, except in the time-step generalization conditions described below. We trained all models to make next-step predictions of all particles' positions and momenta, with a loss function which penalized mean squared error between the true and predicted values.

Fig. \ref{fig:overview}a shows that for models trained and tested on RK4, the \odemodel{} and \hamiltonianmodel{} have lower rollout error than the \deltamodel{}, and even lower than integrating the true Hamiltonian at the same time step. Fig. \ref{fig:overview}b shows that the average energy of the system also stays closer to the initial energy for the \odemodel{} and \hamiltonianmodel{}. The \odemodel{} energy seems to preserve energy better than the \hamiltonianmodel{}, however this is not surprising: as shown below, the \hamiltonianmodel{} matches the \truehamiltonian{} well, and both have imperfect energy conservation when integrated with RK4, which is not symplectic\footnote{Symplectic integrators are designed to have more accurate energy conservation.}. Similar experiments with a third-order symplectic integrator\cite{candy1991symplectic} yielded similar rollout errors for the \hamiltonianmodel{}, with 3 times better energy conservation than the \odemodel{} (See Supp. Sec. \ref{sec:symplectic integrators}). Figs. \ref{fig:overview}c-g (and \href{https://tinyurl.com/hogn-example-trajectories}{\textcolor{blue}{\underline{video link}}}) illustrate the rollout accuracy of the different models for a long 500-step trajectory.

\paragraph{Generalizing to untrained time-steps}

\begin{figure*}[t!]
\begin{center}
    \adjincludegraphics[height=0.138\textheight, trim={{0.00\width} -9 {.02\width} -10},clip=true]{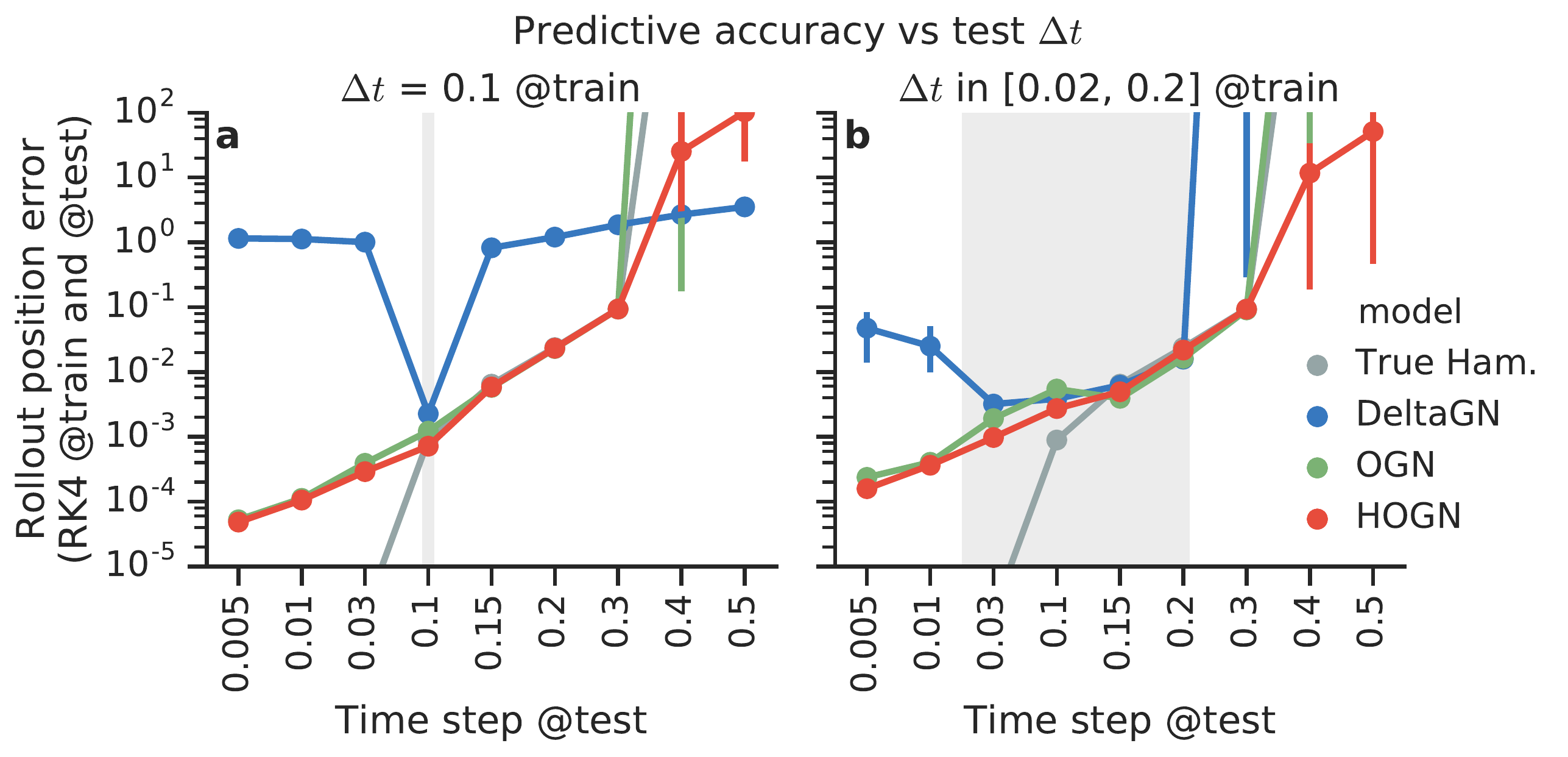}
    \adjincludegraphics[height=0.138\textheight, trim={{0.02\width} 0 {.02\width} 0},clip=true]{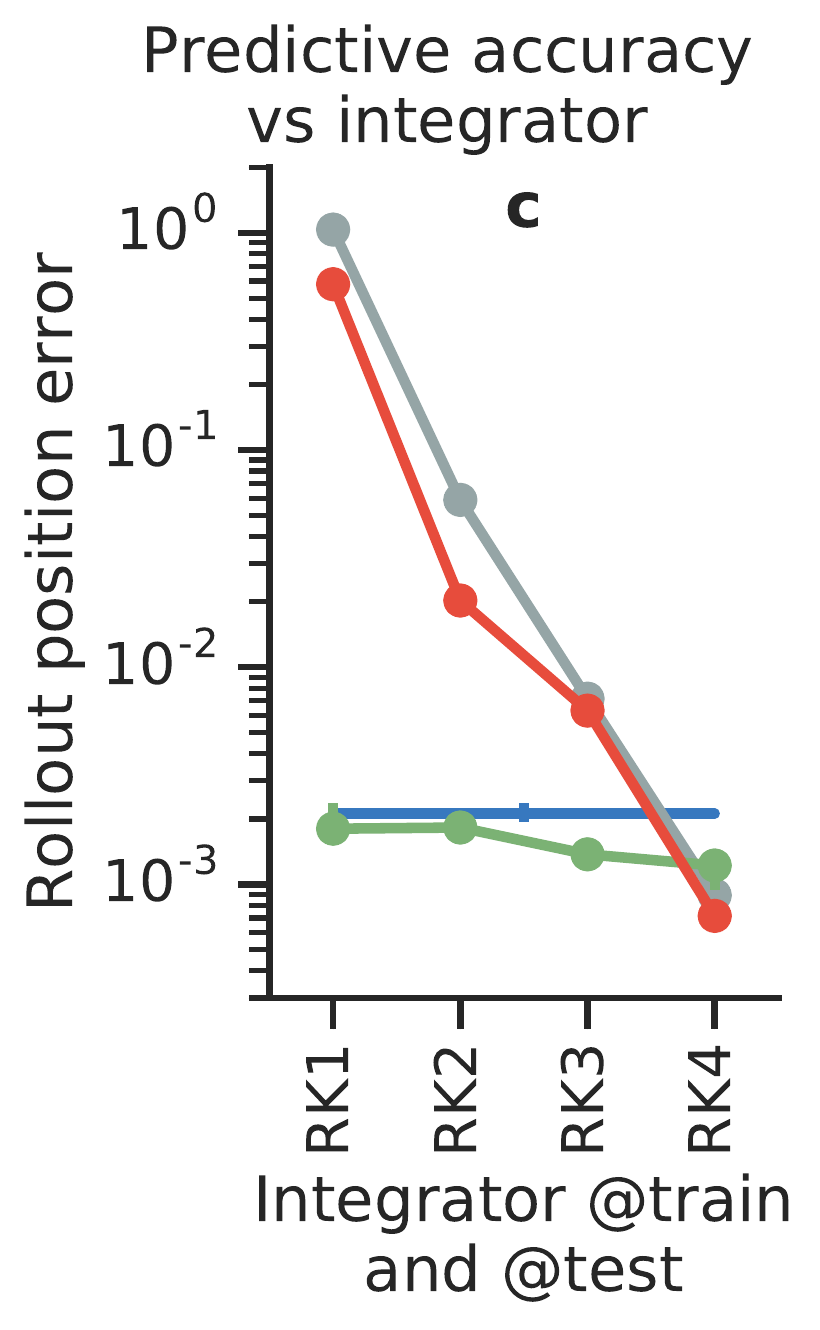}
    \adjincludegraphics[height=0.138\textheight, trim={{0.02\width} 0 {.02\width} 0},clip=true]{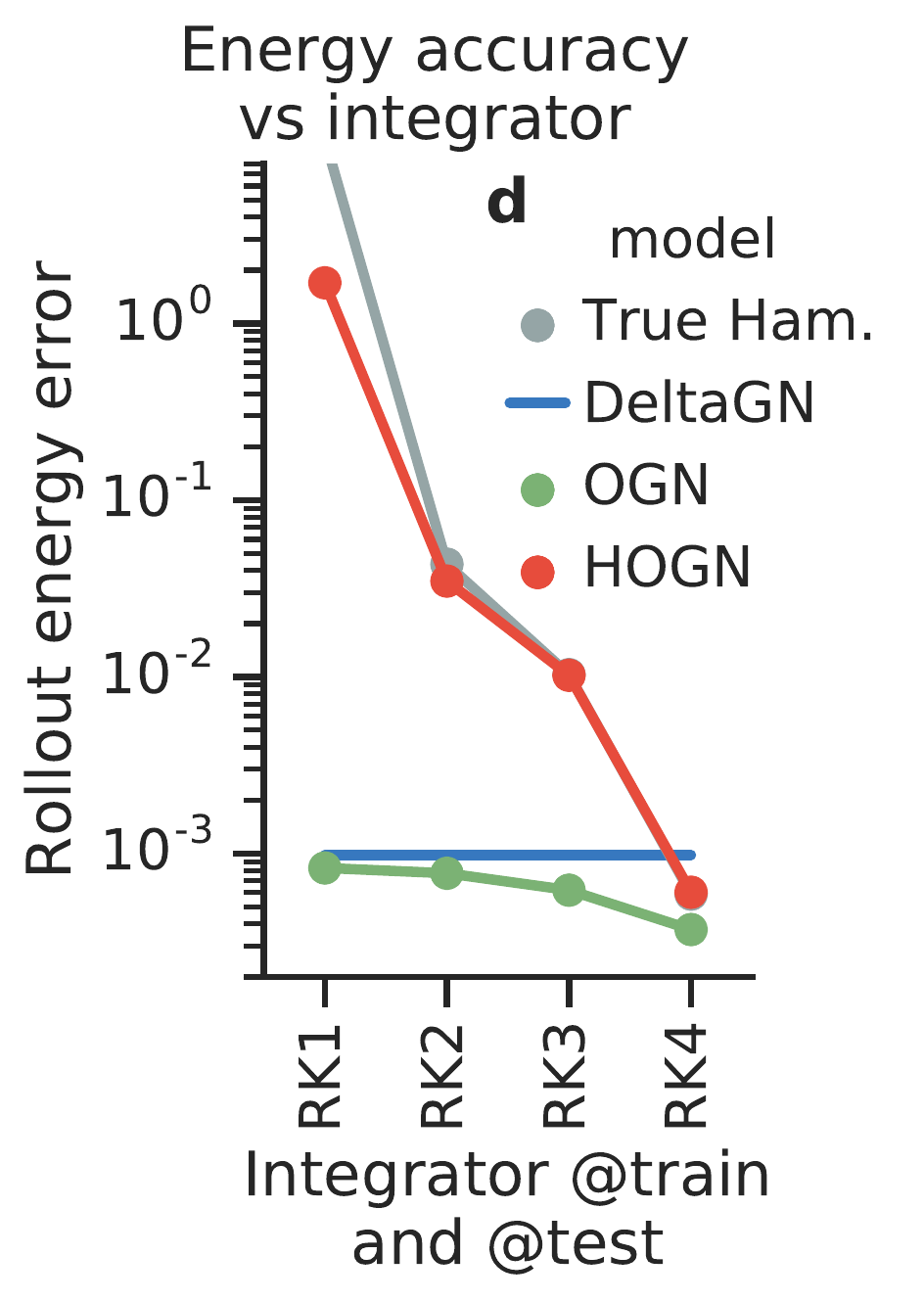}
    \adjincludegraphics[height=0.138\textheight, trim={{0.01\width} 0 {.02\width} 0},clip=true]{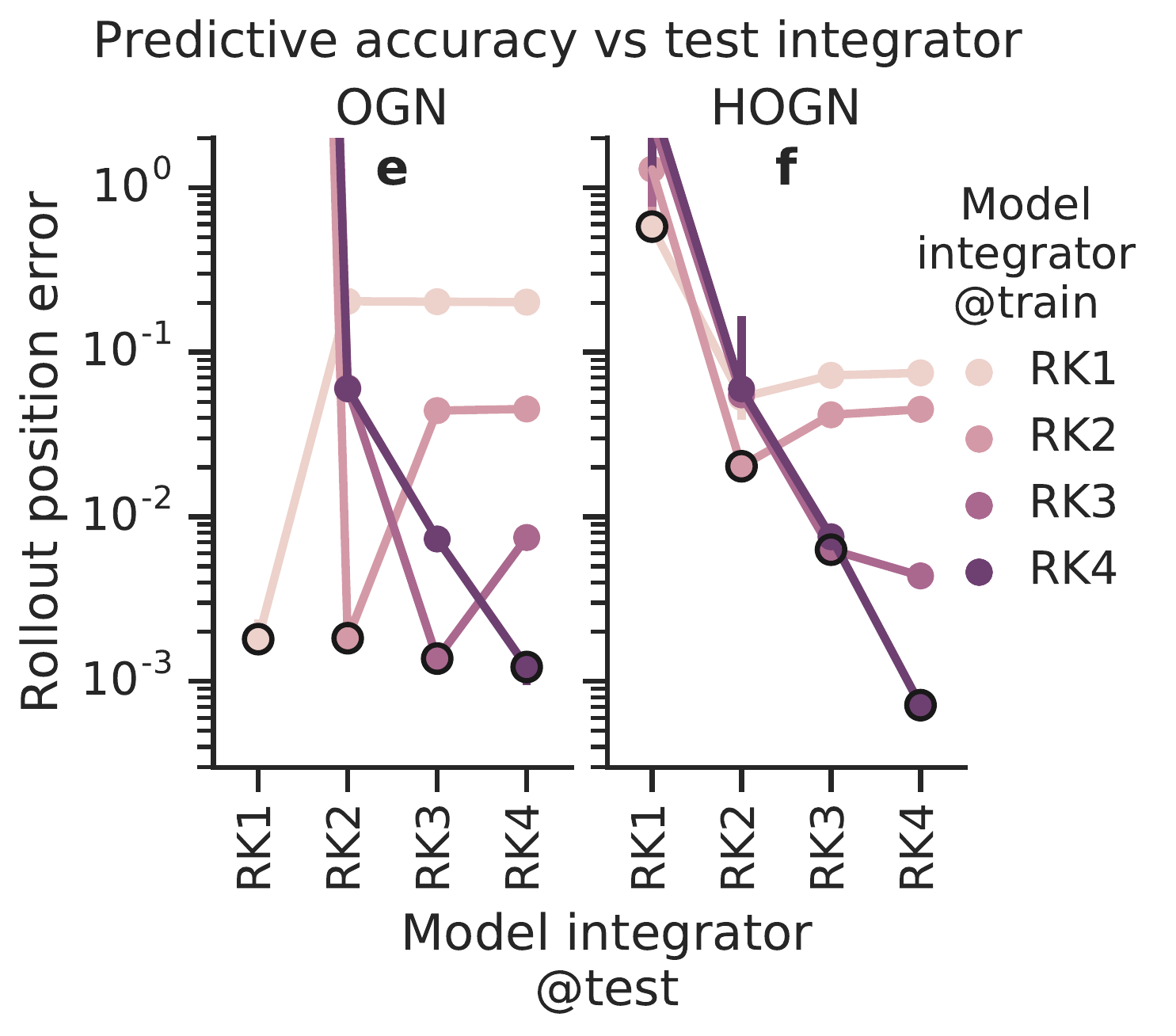}
    \caption{\captionletter{a-b} Time-step size generalization error in 20-step-long trajectories when trained with \captionletter{a} a fixed time-step of 0.1 or \captionletter{b} variable time-steps (0.02-0.2). \captionletter{c-d} Predictive accuracy and energy conservation across models and integrators. \captionletter{e-f} Results when varying the integrator used at test time, where points that share the same train and test integrator are highlighted with black circles.
    }
    \label{fig:generalization}
\end{center}
\end{figure*}

Both the \odemodel{} and \hamiltonianmodel{} exhibit better zero-shot generalization than the \deltamodel{} on time-steps which were never experienced during training. Fig. \ref{fig:generalization}a shows each model's performance when trained on a time-step of $0.1$ and tested on time-steps between $0.005$ and $0.5$. We also trained models on multiple time-steps, from $0.02$ to $0.2$, to study whether that could improve generalization. We varied the time-steps of the ODE models via the integrator, while for the \deltamodel{} we provided the time-step as input to the model. Fig. \ref{fig:generalization}b shows that the \deltamodel{} is on par with the other two models within the training range, but becomes worse outside this range. Note, predictions for long time-steps ($> 0.3$) become very inaccurate for all learned models, as well as the \truehamiltonian, which suggests that the models are determining the dynamics on the basis of features only appropriate for shorter time-steps. Results for other integrators in Supp. Sec. \ref{sec:time_generalization_other_integrators}.

\paragraph{Generalization across integrators}

We varied the integrators used by the model to examine how the quality of the integrator interacts with our different models. Figs. \ref{fig:generalization}c-d show that for lower order integrators (RK1-3), the \hamiltonianmodel{} and \truehamiltonian{} have higher rollout and energy errors than the non-Hamiltonian models (\deltamodel{} and \odemodel{}).
We speculate the non-Hamiltonian models are more accurate because they are not obligated to respect the constraint the Hamiltonian imposes between the $\mathbf{q}$'s and $\mathbf{p}$'s dynamics, and can instead, for instance, learn approximations to the time derivatives which are not consistent with the $\mathbf{q}$ and $\mathbf{p}$ gradient vector fields of any Hamiltonian.

However, Fig. \ref{fig:generalization}e shows that when the \odemodel{} is tested on integrators different from those on which it was trained, its performance is always significantly worse. This indicates that the \odemodel{} is not learning accurate $\left(\mathbf{\dot q}, \mathbf{\dot p}\right)$, but rather some other function, which interfaces with the training-time integrator to produce accurate trajectories, but is inappropriate for other integrators. By contrast, Fig. \ref{fig:generalization}f shows that the \hamiltonianmodel{} model usually generalizes to greater accuracy when used with higher order integrators at test time, and specifically when trained with RK4 (dark purple line in Fig. \ref{fig:generalization}f), perfectly matches the behavior of the true Hamiltonian (grey line in Fig. \ref{fig:generalization}c). 

\citet{greydanus2019hamiltonian}, in their Hamiltonian Neural Networks, use finite differences to approximate the derivatives of $\mathbf{q}$ and $\mathbf{p}$ when not available analytically, use those approximated values during training to supervise the Hamiltonian gradients, and then generate test trajectories with an RK4 integrator. In our experiments this is analogous to training the \hamiltonianmodel{} using RK1 and testing on RK4, and indeed, our findings are generally consistent with theirs, but also indicate that directly learning through higher order integrators is generally better.

\section{Discussion and future work}

We incorporated two physically informed inductive biases---ODE integrators and Hamiltonian mechanics---into graph networks for learning simulation, and found they could improve performance, energy, and zero-shot time-step generalization. We also analyzed the impact of varying the ODE integrator independently between training and test, and found that the Hamiltonian approaches benefit most from training with an RK4 integrator, while non-Hamiltonian ones can learn more accurate predictions when trained on lower order integrators. However the \hamiltonianmodel{} generalizes better across integrators, and approximates the \truehamiltonian{} best in terms of performance and energy accuracy. 

Similar to the graph network's inductive bias for representing complex systems as nodes interacting via edges, the ODE integrator and Hamiltonian inductive biases are informed by physics, but are not specific to physics. Our approach may be applicable to many complex multi-entity systems which can be modeled with ODEs or have some notion of canonical position and momentum coordinates. Moreover, there are other physically informed inductive biases to explore in future work, such as imposing separable potential and kinetic energy terms, Lagrangian mechanics, time-reversibility, or entropic constraints.

\section*{Acknowledgments}

We acknowledge Irina Higgins, Danilo Rezende, Razvan Pascanu, Shirley Ho and Miles Cranmer for helpful discussions and comments on the manuscript.

\bibliography{main}
\bibliographystyle{unsrtnat}

\appendix
\newpage

\renewcommand\thefigure{\thesection.\arabic{figure}}
\renewcommand\thetable{\thesection.\arabic{table}}

\setcounter{figure}{0}
\setcounter{table}{0}

\section{Data generation} 

\subsection{Initial conditions}

Each particle $i$ had initial conditions (all in figures in international system of units) drawn from independent uniform distributions:

\begin{itemize}
\item Mass $m^i \in [0.1, 1]$
\item Spring constant $k^{i} \in [0.5, 1]$
\item Initial position $\textbf{q}^i_0 \in [-1, 1]^2$
\item Initial velocity $\textbf{v}^i_0 \in [-3, 3]^2$, $\textbf{p}^i_0 = m^i\textbf{v}^i_0$
\end{itemize}

\subsection{Physics simulator}

Our dataset consists of simulations for particle-spring systems where particle $j$ exerts a force on particle $i$ (Hooke's force) $\mathbf{F}^{ij} = -k^{ij}\cdot(\mathbf{q}^i - \mathbf{q}^j)$
where $k^{ij}$ is the spring constant, calculated as $k^{ij} = k^i\cdot k^j$. 

Trajectories were generated using RK4 with a generation time step of 0.005s, with trajectories of up to 4s in length.

\subsection{Dataset}

The datasets were obtained by subsampling the trajectories generated with the physics simulator at different time intervals.

We generated data for systems with between 2 and 15 particles, although we only used data with 4, 5, 6, 8 and 9 particles for training.

The training dataset consisted of 10000 one-step pairs of states (for each number of particles) separated by either a fixed time step of {0.1 s}, or by a random time step drawn uniformly from the {$[0.02, 0.2]$ s} interval and rounded to the nearest exact multiple of the generation time-step. In the latter case, to avoid having discrete values of the time step, instead of using a fixed generation time-step of {0.005 s}, we used random values form the interval {0.005 s $\pm$ 10\%}. Each pair was sampled from a different randomly generated trajectory. Additionally we generated 1000 validation and 1000 test pairs for each number of particles.

We also stored validation and test trajectories of length 20 sampled at 0.005, 0.01, 0.03, 0.1, 0.15, 0.2, 0.3, 0.4, 0.5 seconds (1000 trajectories for each number of particles and time step). All results presented in the paper correspond to the full test dataset of 20-step test trajectories for systems with 4, 5, 6, 8 and 9 particles.

\section{Graph Network} 
\label{supp_sec:gn_parameters}

Each of the MLPs used on the edge model, node model and global model of the graph network, have output sizes of [64, 64], and \emph{softplus} activations after each layer including the last one. The graph network always uses \emph{sum} as the aggregation function for edges and nodes.

The output of the network is obtained by applying an additional non-activated linear layer with the desired output size to the output globals (\hamiltonianmodel{}, output size of 1 to represent a single scalar value per graph) or to the output nodes (\odemodel{} and \deltamodel{}, output size of 4 to represent values associated to 4 canonical coordinates for each node/particle).

We chose \emph{softplus} because \emph{relu} activation did not work well with the Hamiltonian model. We hypothesize this is a consequence of Eq. \ref{eq:network_output_derivatives}: a function approximator with \emph{relu} activation is a piecewise linear function (linear almost everywhere), so after taking the derivatives of the output of the network with respect to its inputs, it becomes a function that is constant almost everywhere, and the gradient-based optimization process struggles to optimize it. This is consistent with \cite{greydanus2019hamiltonian} where they find better results for \emph{tanh} activation than for \emph{relu} activation. In our case \emph{softplus} seemed to produce better results than \emph{tanh}.

We always remove the mean position of the objects from the inputs to the GraphNetwork. This does not prevent the models from predicting correct absolute positions since all dynamics are translation invariant and models (Including RK integrators) only predict relative differences with respect to the previous state. This allows the model to produce correct predictions even when particles have travelled far away from their initial positions.

Also, because the ground truth Hamiltonian in spring systems is time-independent (it is a conservative system), we omit feeding absolute time to the GraphNetworks to avoid unnecessary dependencies.

\section{Training details} 
\label{supp_sec:training_parameters}

We used TensorFlow and the DeepMind Graph Nets library to build our  neural networks, calculate the analytic gradients of the Hamiltonian, and train our models. We trained with a batch size of 100, for a million training steps. We used AdamOptimizer to minimize the loss.

Due to the different nature of the models (which may require different learning rates) we ran each of the models with 13 initial learning rates uniformly spanning an interval between $10^{-1}$ and $10^{-4}$ in log scale. Learning rate was decayed exponentially at a rate of 0.1 every $2\cdot 10^5$, with a lower limit of $10^{-7}$.

For each model, we report median values and min-max range for the 4 learning rates with the smallest rollout error. In practice the variance across seeds given the same learning rate was negligible (except on generalization results). 

Rollout error is defined as the RMS position error averaged across all examples, dimensions, particles, and sequence axis. Energy error is calculated as the RMS of the deviation between the mean energy of a trajectory and the initial energy (normalized by the initial energy, to yield relative errors), averaged across all examples.

\section{Additional results} 

\subsection{Symplectic integrators}
\label{sec:symplectic integrators}

We attempted using symplectic integrators of first order (Symplectic Euler, S1), second order (Verlet Integrator, S2) and third order (S3) with coefficients as described in \cite{candy1991symplectic}\footnote{We also tried a fourth order symplectic integrator, as described in \cite{candy1991symplectic}, however it did not improve results, possibly due to the linear nature of the Hooke's force\cite{symplectic34}.}. It is worth noting that these integrators are only symplectic for separable Hamiltonians $\mathcal{H}(\mathbf{q}, \mathbf{p}) = T(\mathbf{p}) + V(\mathbf{q})$, or equivalently, for systems for which $\mathbf{\dot p} = f_{\mathbf{\dot p}}(\mathbf{q})$ and $\mathbf{\dot q} = f_{\mathbf{\dot q}}(\mathbf{p})$. However, we do not impose this constraint, neither to the \hamiltonianmodel{} (by adding the scalar output of two independent networks, with $\mathbf{p}$ and  $\mathbf{q}$ inputs respectively) nor the \odemodel{} (by only feeding $\mathbf{q}$ to the network that outputs $\mathbf{\dot p}$ and vice versa), so there are not any guarantees that the learned models will correctly preserve energy for any Hamiltonian.

\begin{figure*}[t!]
\begin{center}
    \adjincludegraphics[height=0.15\textheight, trim={{0.00\width} -9 {.02\width} -10},clip=true]{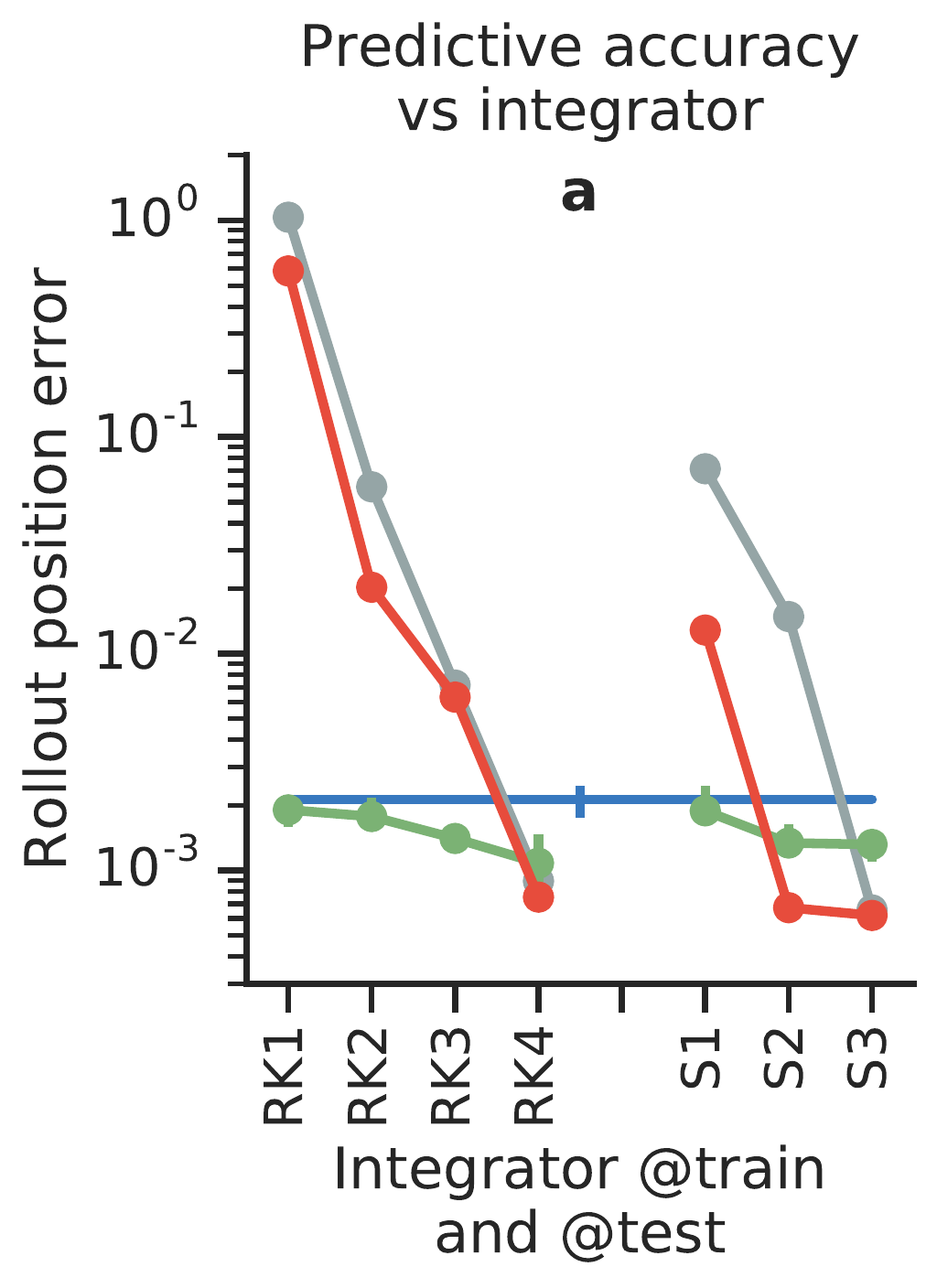}
    \adjincludegraphics[height=0.15\textheight, trim={{0.00\width} -9 {.02\width} -10},clip=true]{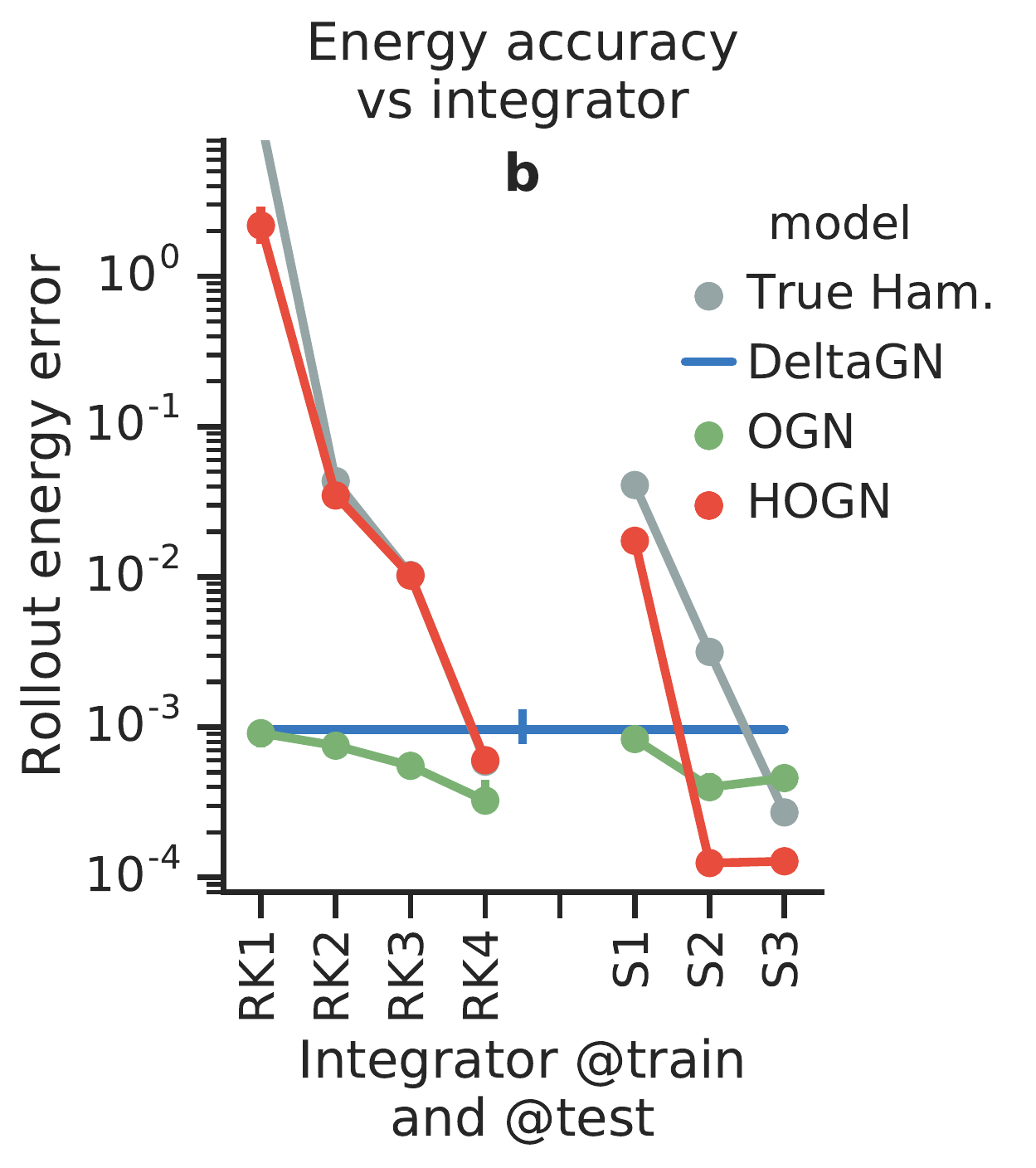}
    \caption{\captionletter{a-b} Predictive accuracy and energy conservation across models and integrators including symplectic integrators (analogous to Fig. \ref{fig:generalization}c-d).
    }
    \label{fig:overall_results_symplectic}
\end{center}
\end{figure*}

\begin{figure*}[t!]
\begin{center}
    \adjincludegraphics[height=0.15\textheight, trim={{0.00\width} -9 {.02\width} -10},clip=true]{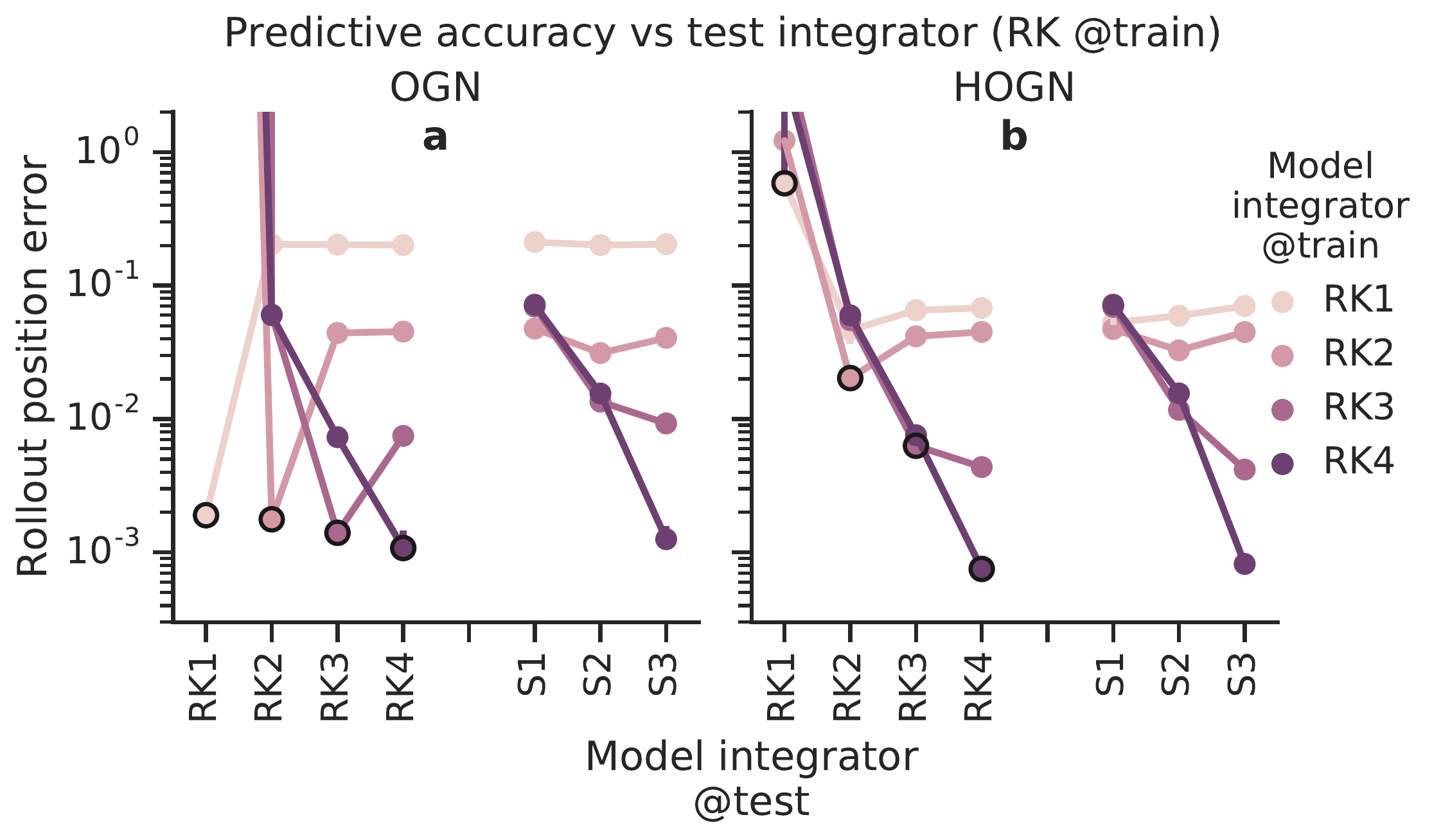}
    \adjincludegraphics[height=0.15\textheight, trim={{0.00\width} -9 {.02\width} -10},clip=true]{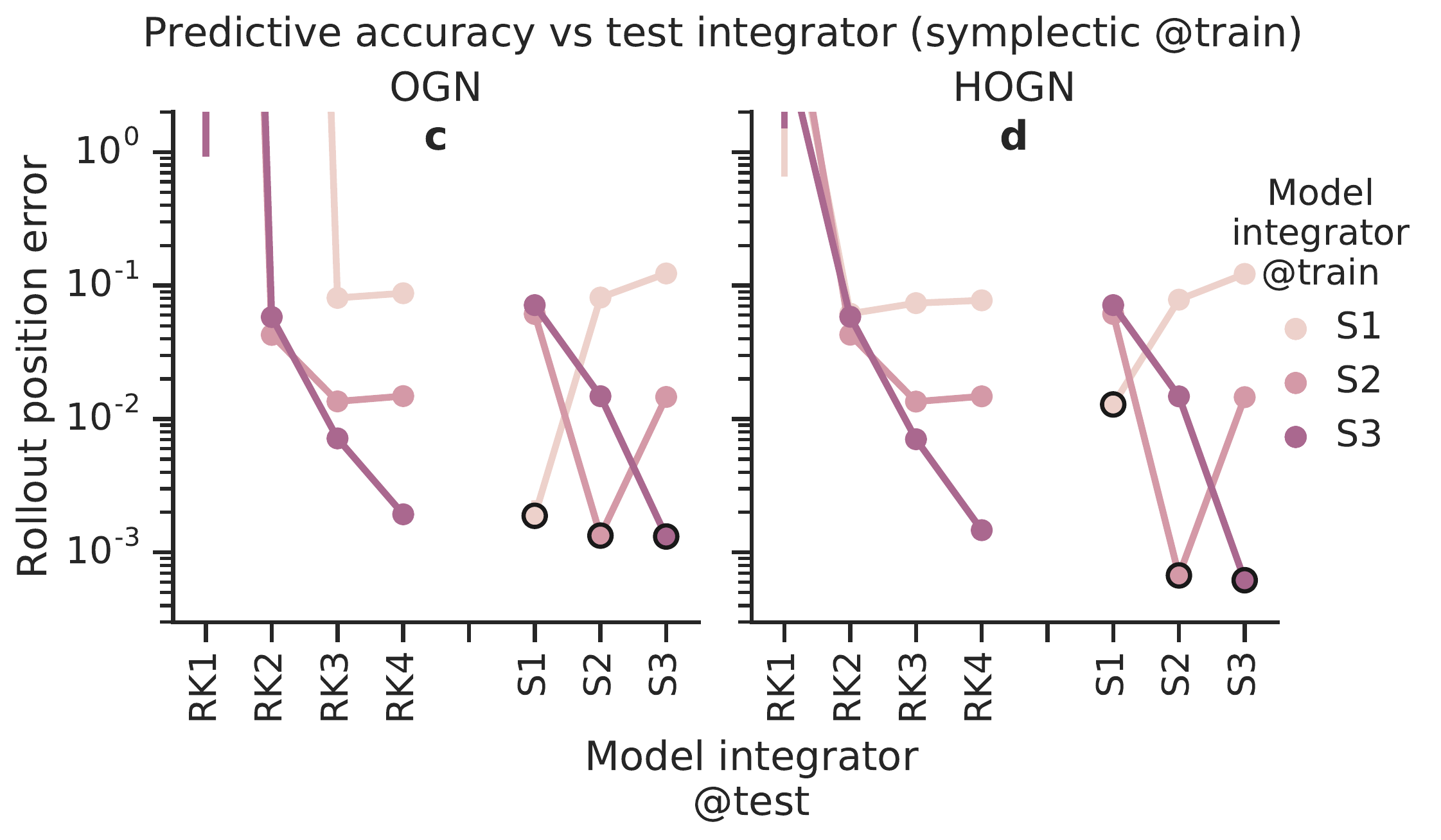}
    \caption{\captionletter{a-d} Predictive accuracy when varying the integrator used at test time, where points that share the same train and test integrator are highlighted with black circles (analogous to Fig. \ref{fig:generalization}e-f).
    }
    \label{fig:integrator_generalization_symplectic_position}
\end{center}
\end{figure*}

Fig. \ref{fig:overall_results_symplectic}a shows the predictive accuracy including results for models trained with symplectic integrators. Contrary to the RK integrators, in this case the \hamiltonianmodel{} seems to be able to produce lower errors than the \truehamiltonian{} even for low order integrators (S1 and S2), in a similar way to how the \odemodel{} was able to produce very low errors for low order RK integrators. While this seems like an advantage of using symplectic integrators it also indicates that the \hamiltonianmodel{} model is not really learning something close to the \truehamiltonian{} in these cases, and, as we will see in the next section, it will cause worse generalization to unseen test time-steps. Similarly, Fig. \ref{fig:integrator_generalization_symplectic_position}d shows that the models trained with the S1, S2 integrators do not generalize at all to higher order symplectic integrators, and only the model trained with S3 shows good generalization across all integrators that is comparable to the generalization across integrators of the model trained with RK4 (Fig. \ref{fig:integrator_generalization_symplectic_position}b) and the behavior of the \truehamiltonian{} (Fig. \ref{fig:overall_results_symplectic}a).

\begin{figure*}[t!]
\begin{center}
    \adjincludegraphics[height=0.15\textheight, trim={{0.00\width} -9 {.02\width} -10},clip=true]{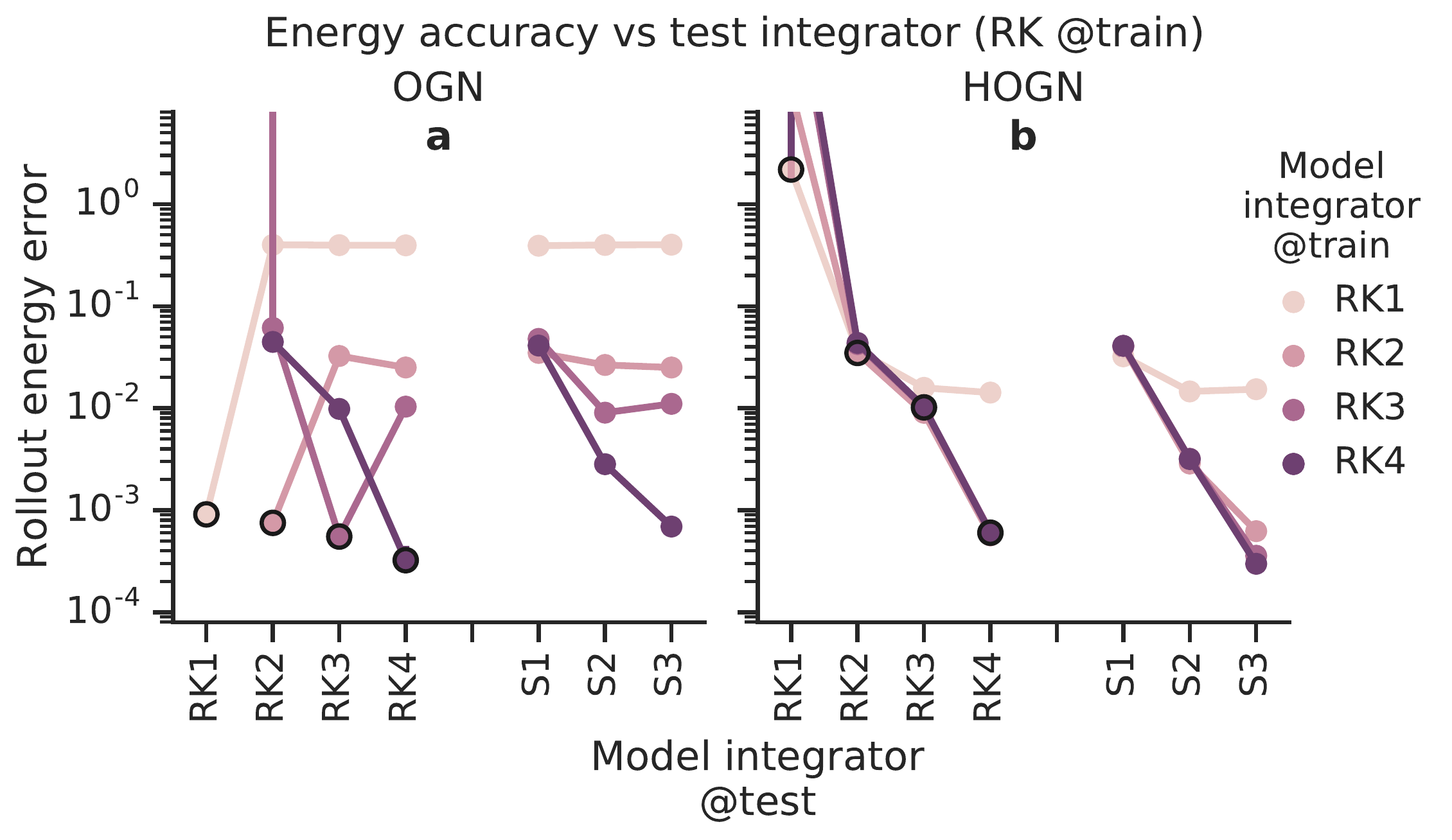}
    \adjincludegraphics[height=0.15\textheight, trim={{0.00\width} -9 {.02\width} -10},clip=true]{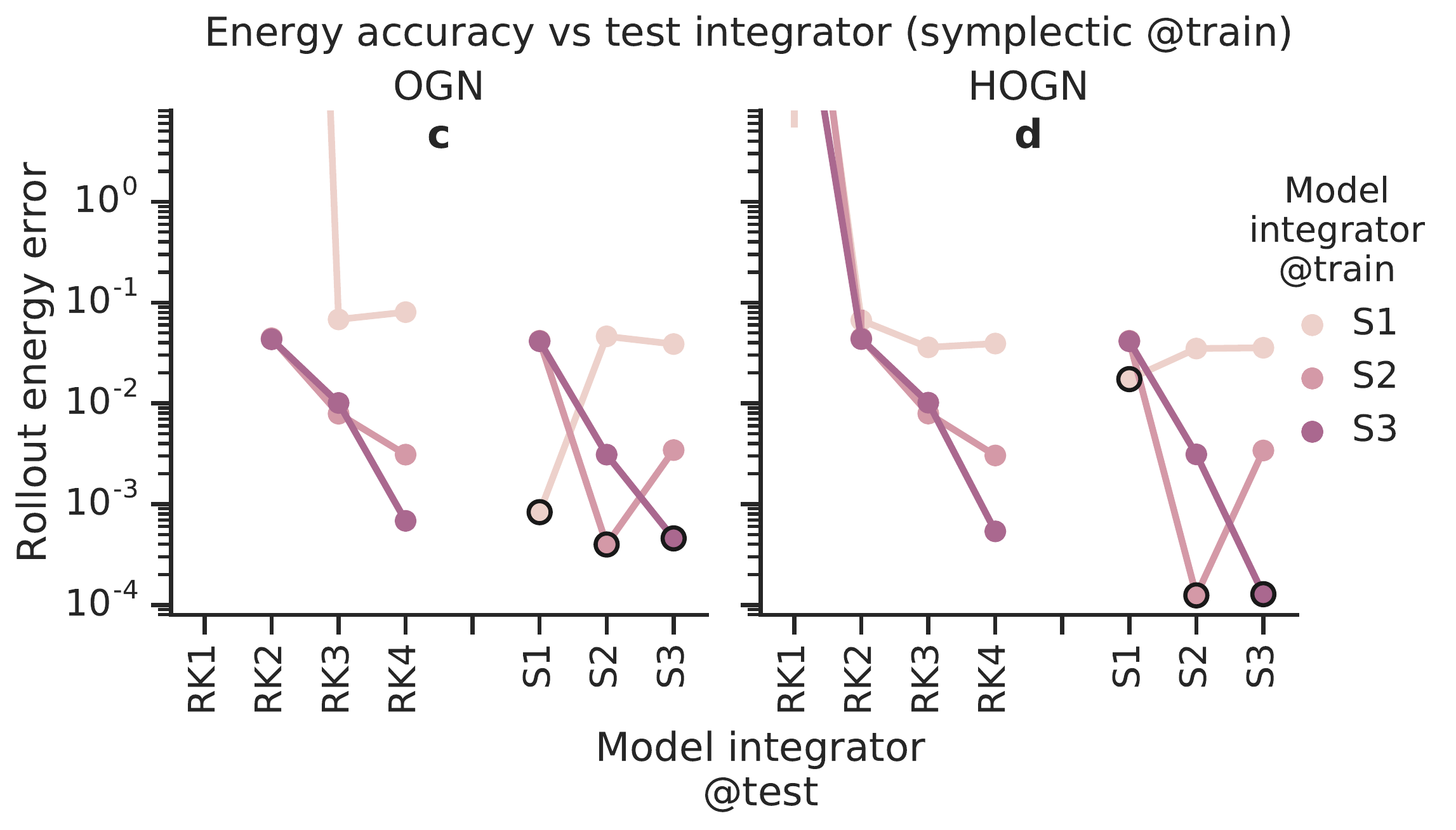}
    \caption{\captionletter{a-d} Energy accuracy when varying the integrator used at test time, where points that share the same train and test integrator are highlighted with black circles (analogous to Fig. \ref{fig:integrator_generalization_symplectic_position}).
    }
    \label{fig:integrator_generalization_symplectic_energy}
\end{center}
\end{figure*}

We speculate these differences in accuracy between training the \hamiltonianmodel{} with the RK integrators and the symplectic integrators are related to fundamental differences between the algorithms behind the two types of integrators. Firstly while the output of RK integrators is usually a weighted average of the results obtained across several internal iterations for intermediate time-steps, the symplectic integrators feed each internal iteration with just the output of the previous internal iteration, and output just the result of the very last internal iteration. This means that the trained models can learn to encode additional information in the outputs of the intermediate internal iterations, which may allow the neural networks to distinguish whether the gradients are being evaluated for the first, the second, etc. or the final internal iteration. Secondly, in RK integrators, the derivatives of the position and momentum are always evaluated at the same point in each internal iteration, but in the symplectic integrators the derivatives of the position and the momentum are evaluated at different points, in alternating fashion. This has two consequences, the first consequence is that an $n$-th order symplectic integrator performs $2n$ network evaluations, while a $n$-th order RK integrator performs only $n$ network evaluations, as it shares the evaluations of the position and momentum gradients at the same point. The additional network evaluations may explain why apparently lower order symplectic integrators, when used with learned models, can produce lower errors (e.g. S2 vs RK2). The second consequence is that, if as speculated before, the model can identify which calls corresponds to which internal iterations, it may also identify when it is being evaluated to produce gradients for the momentum or for the position. This would allow the model to get around the Hamiltonian constraint (that used to force it to produce a position and momentum vector field of gradients consistent with a common Hamiltonian), and allow it to start behaving more similarly to the \odemodel{}, where the derivatives of the position and momentum can be fully independent.

With respect to energy conservation, Fig. \ref{fig:overall_results_symplectic}b, shows that models trained with higher order symplectic integrators (S1 and S2) preserve energy much better than the RK integrators, and, as mentioned in the main text, allow the \hamiltonianmodel{} to preserve energy much better than the \odemodel{}. It is also worth noting, that the \hamiltonianmodel{} trained with RK2, RK3 and RK4 integrators also improves in energy conservation when integrated with symplectic integrators (Fig. \ref{fig:integrator_generalization_symplectic_energy}b), confirming that the models trained with high order RK integrators learn something very similar to to the ground truth Hamiltonian that even generalizes to a different family of integrators.

\subsection{Time generalization for other integrators}
\label{sec:time_generalization_other_integrators}

\begin{figure*}[t!]
\begin{center}
    \adjincludegraphics[height=0.138\textheight, trim={{0.00\width} -9 {.02\width} -10},clip=true]{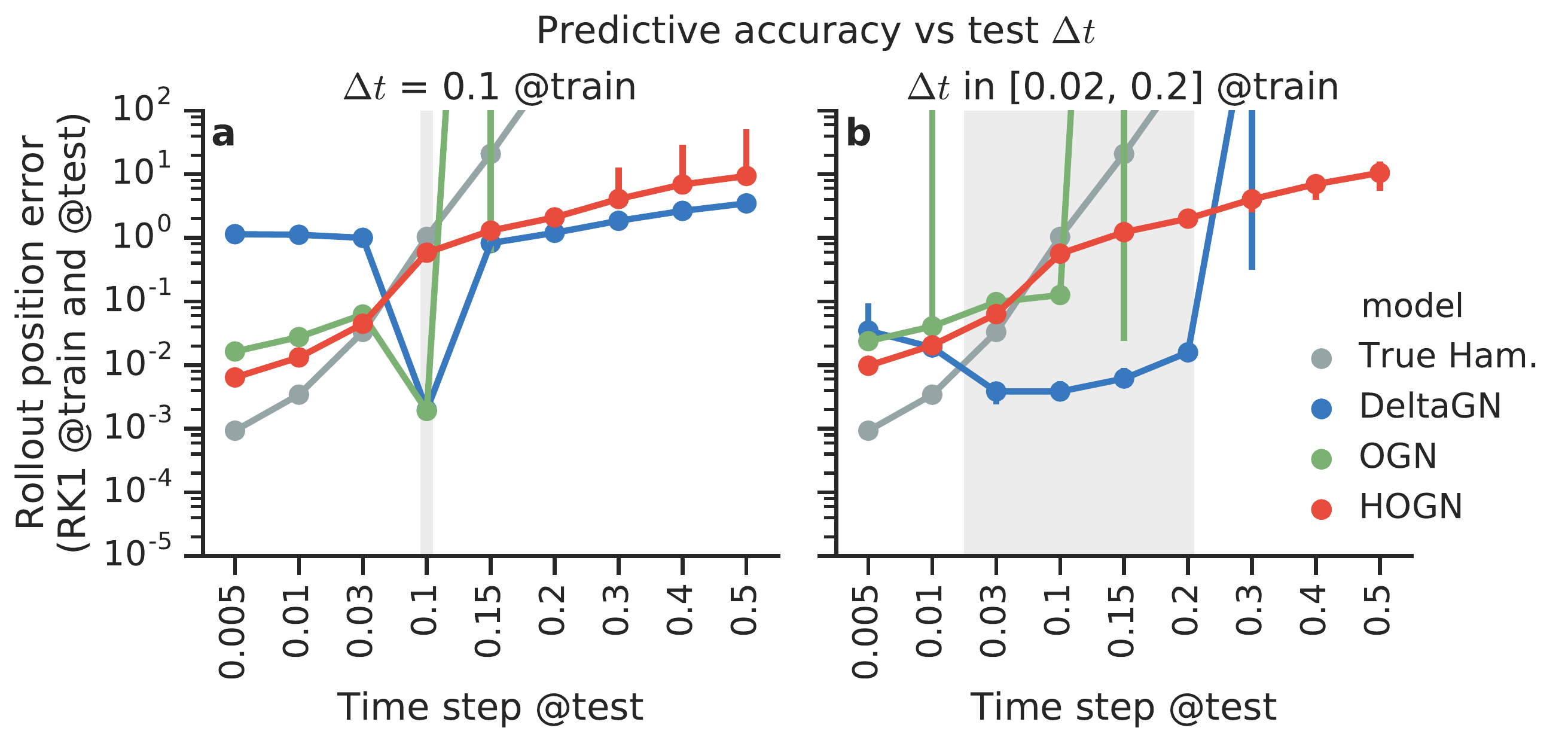}
    \caption{\captionletter{a-b} Time generalization for RK1 integrator (analogous to Fig. \ref{fig:generalization}a-b).
    \label{fig:time_generalization_rk1}
    }
\end{center}
\end{figure*}

\begin{figure*}[t!]
\begin{center}
    \adjincludegraphics[height=0.138\textheight, trim={{0.00\width} -9 {.02\width} -10},clip=true]{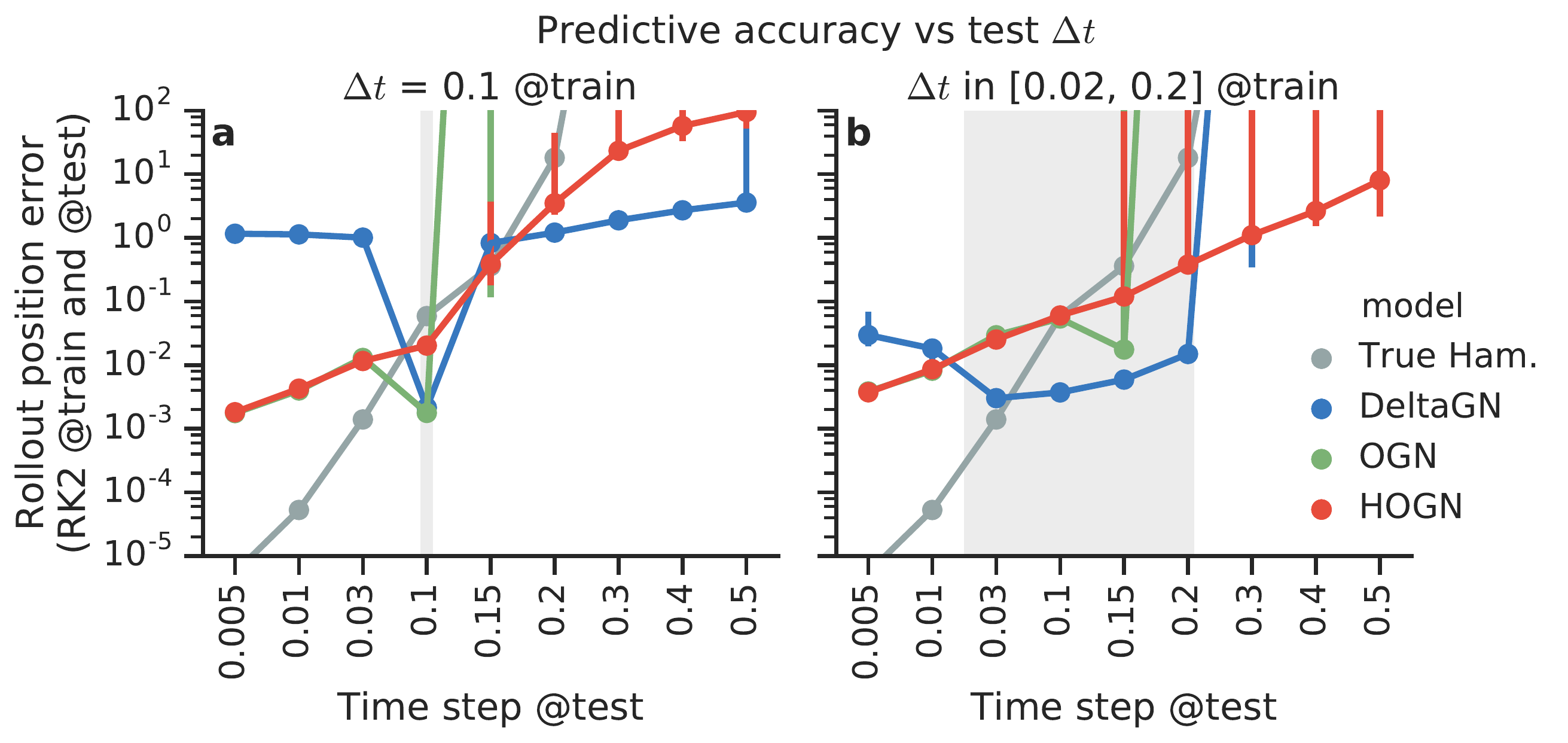}
    \caption{\captionletter{a-b} Time generalization for RK2 integrator (analogous to Fig. \ref{fig:generalization}a-b).
    \label{fig:time_generalization_rk2}
    }
\end{center}
\end{figure*}

\begin{figure*}[t!]
\begin{center}
    \adjincludegraphics[height=0.138\textheight, trim={{0.00\width} -9 {.02\width} -10},clip=true]{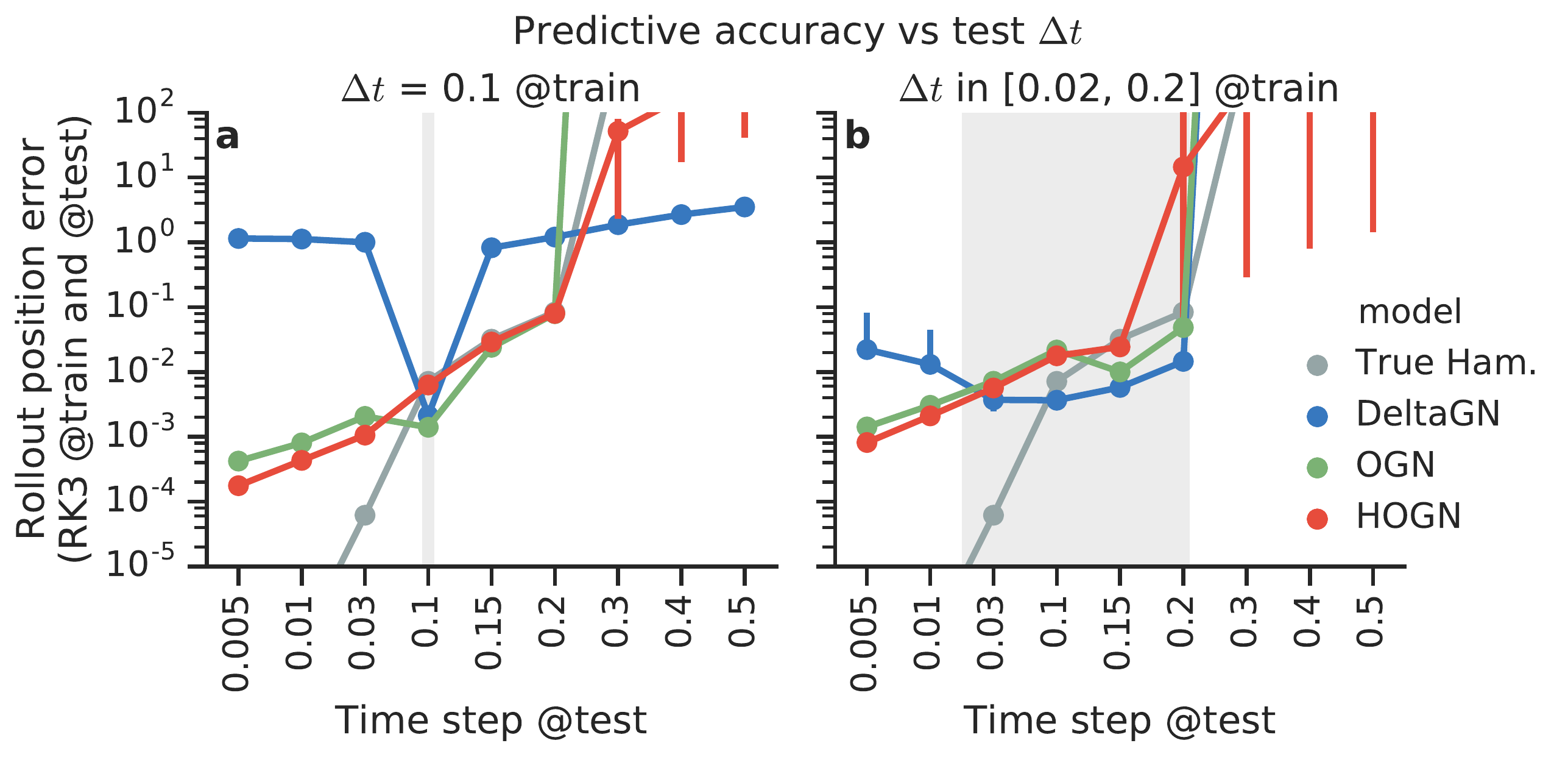}
    \caption{\captionletter{a-b} Time generalization for RK3 integrator (analogous to Fig. \ref{fig:generalization}a-b).
    \label{fig:time_generalization_rk3}
    }
\end{center}
\end{figure*}

Figs. \ref{fig:time_generalization_rk1}, \ref{fig:time_generalization_rk2}, and \ref{fig:time_generalization_rk3} show that for RK integrators of lower order (RK1-RK3) the \odemodel{} overfits heavily to the time-step used during training, however, the \hamiltonianmodel{} presents consistent generalization to time-steps not seen during training, yielding approximately similar errors. This again is evidence that the \hamiltonianmodel{} learns something more consistent with a Hamiltonian, and that in fact follows more closely the \truehamiltonian{} than the \odemodel{}.

\begin{figure*}[t!]
\begin{center}
    \adjincludegraphics[height=0.138\textheight, trim={{0.00\width} -9 {.02\width} -10},clip=true]{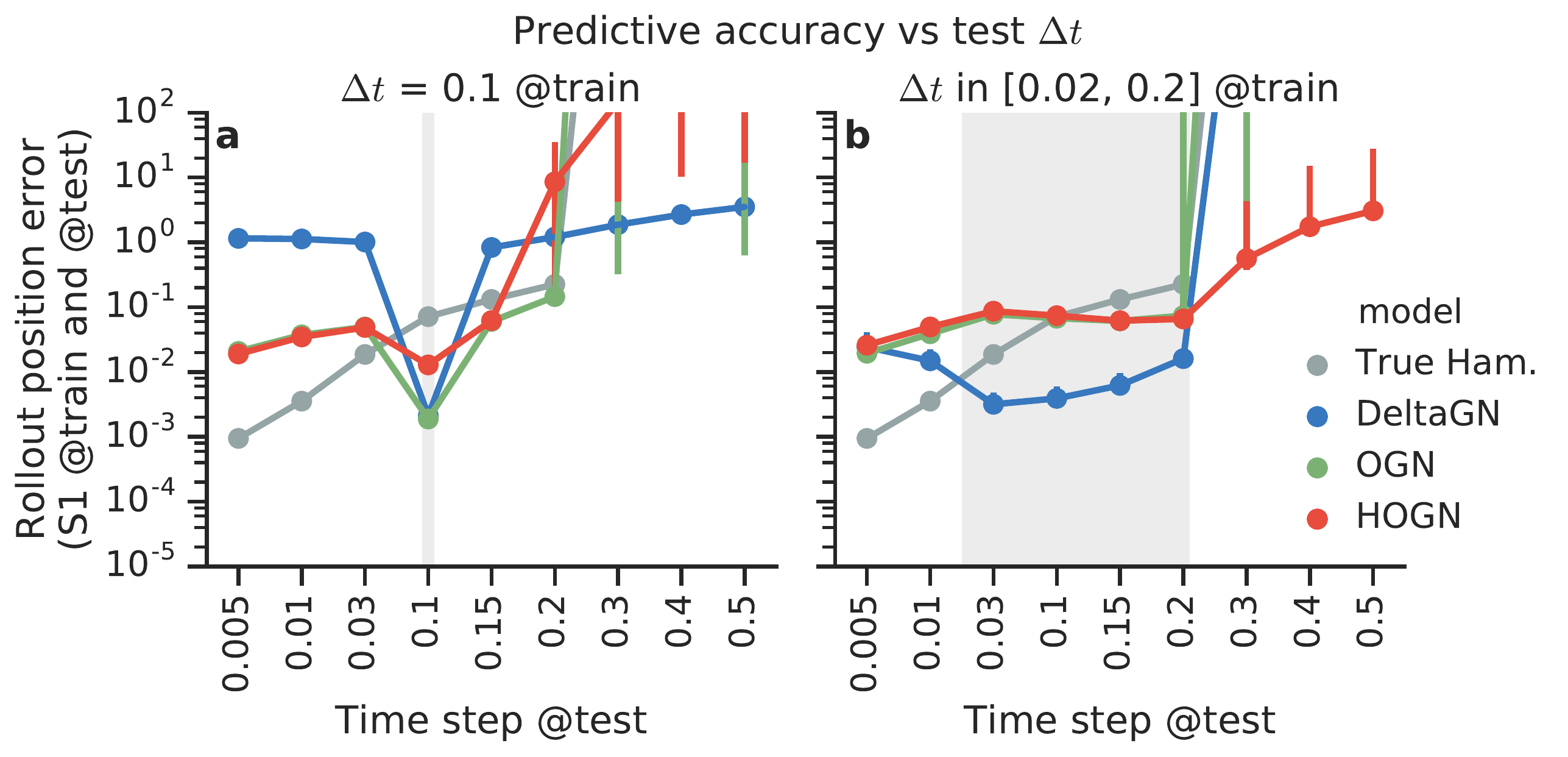}
    \caption{\captionletter{a-b} Time generalization for S1 integrator (analogous to Fig. \ref{fig:generalization}a-b).
    \label{fig:time_generalization_s1}
    }
\end{center}
\end{figure*}

\begin{figure*}[t!]
\begin{center}
    \adjincludegraphics[height=0.138\textheight, trim={{0.00\width} -9 {.02\width} -10},clip=true]{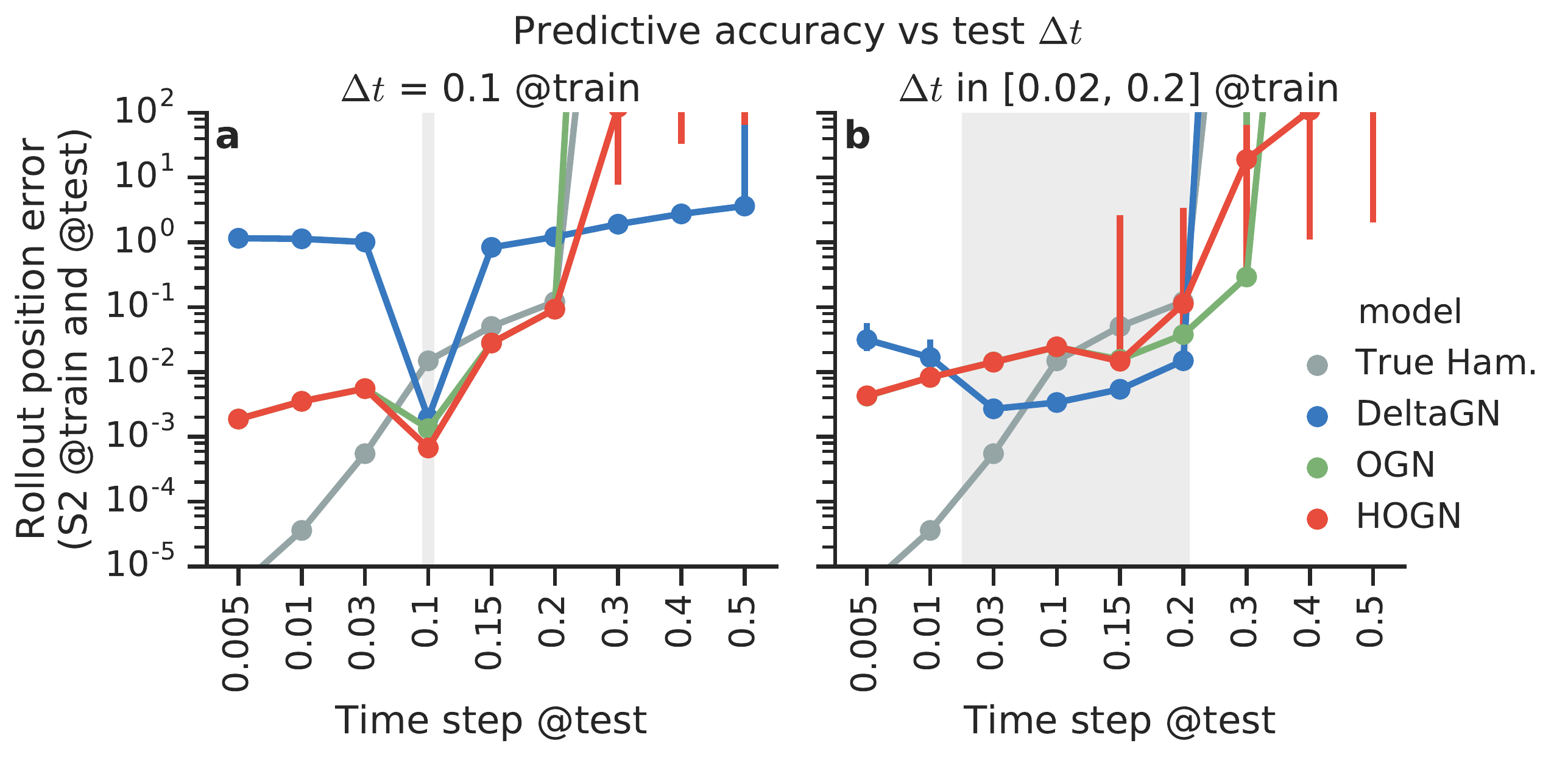}
    \caption{\captionletter{a-b} Time generalization for S2 integrator (analogous to Fig. \ref{fig:generalization}a-b).
    \label{fig:time_generalization_s2}
    }
\end{center}
\end{figure*}

\begin{figure*}[t!]
\begin{center}
    \adjincludegraphics[height=0.138\textheight, trim={{0.00\width} -9 {.02\width} -10},clip=true]{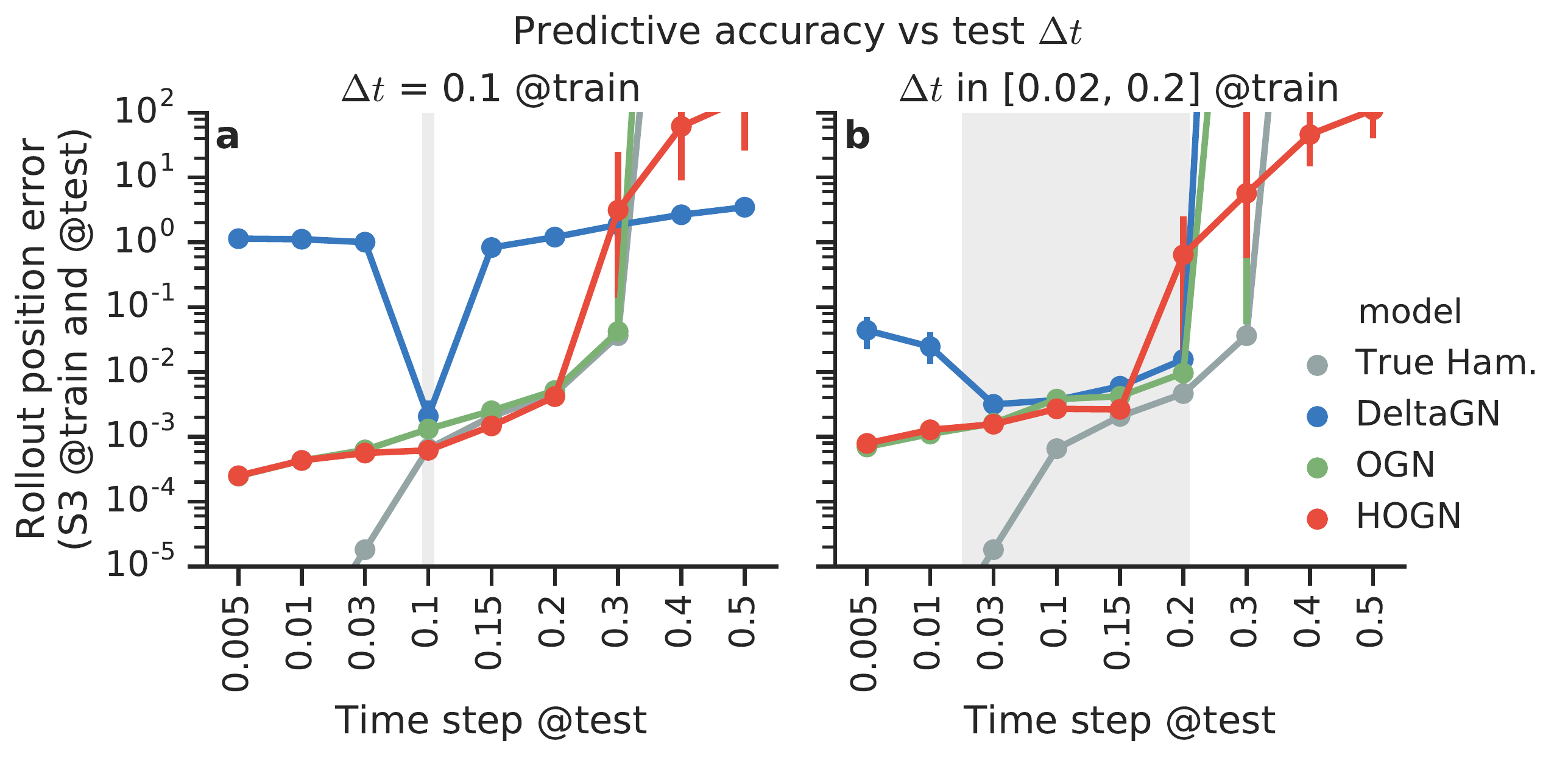}
    \caption{\captionletter{a-b} Time generalization for S3 integrator (analogous to Fig. \ref{fig:generalization}a-b).
    \label{fig:time_generalization_s3}
    }
\end{center}
\end{figure*}

In the case of symplectic integrators, we observe that even the \hamiltonianmodel{} overfits to the time-step used during training when using first order (S1) and second order (S2) integrators (Figs. \ref{fig:time_generalization_s1} and \ref{fig:time_generalization_s2}), and starts generalizing well only with the third order (S3) integrator (Fig. \ref{fig:time_generalization_s3}), although even for S3 the generalization to longer (> 0.2) time-steps is comparatively worse than with RK4 (Fig. \ref{fig:generalization}a-b). This is consistent with the previous discussion (Section \ref{sec:symplectic integrators}) in that the \hamiltonianmodel{} is not really learning an accurate Hamiltonian when trained with a symplectic integrator.

\end{document}